\documentclass{article}

\usepackage{arxiv}

\usepackage[utf8]{inputenc} 
\usepackage[T1]{fontenc}    
\usepackage{hyperref}       
\usepackage{url}            
\usepackage{booktabs}       
\usepackage{amsfonts}       
\usepackage{nicefrac}       
\usepackage{microtype}      
\usepackage{lipsum}
\usepackage{graphicx}
\graphicspath{ {./images/} }

\title{Supervised Machine Learning for Breast Cancer Risk Factors Analysis and Survival Prediction}

\author{
 Khaoula Chtouki \\
  Meridian Team, LYRICA Laboratory\\
  School of Information Sciences\\
  Rabat, Morocco \\
  \texttt{khaoula.chtouki@esi.ac.ma} \\
   \And
 Maryem Rhanoui \\
  Meridian Team, LYRICA Laboratory\\
  School of Information Sciences\\
  Rabat, Morocco\\
  \texttt{mrhanoui@esi.ac.ma} \\
  \And
  \And
 Mounia Mikram \\
    Meridian Team, LYRICA Laboratory\\
    School of Information Sciences\\
    Rabat, Morocco\\
    \texttt{mmikram@esi.ac.ma} \\
  \And
Siham Yousfi \\
 Meridian Team, LYRICA Laboratory\\
 School of Information Sciences\\
 Rabat, Morocco\\
 \texttt{syousfi@esi.ac.ma} \\
  \And
 Kamelia Amazian \\
  Laboratoire Pathologie Humaine, Biomedecine et Environnement\\ 
  Faculty of Medicine and Pharmacy\\
  Fez, Morocco \\
  \texttt{k.amazian@ispitsfes.ac.ma} \\
}

\begin{document}
\maketitle
\begin{abstract}
The choice of the most effective treatment may eventually be influenced by breast cancer survival prediction. To predict the chances of a patient surviving, a variety of techniques were employed, such as statistical, machine learning, and deep learning models. In the current study, 1904 patient records from the METABRIC dataset were utilized to predict a 5-year breast cancer survival using a machine learning approach. In this study, we compare the outcomes of seven classification models to evaluate how well they perform using the following metrics: recall, AUC, confusion matrix, accuracy, precision, false positive rate, and true positive rate. The findings demonstrate that the classifiers for Logistic Regression (LR), Support Vector Machines (SVM), Decision Tree (DT), Random Forest (RD), Extremely Randomized Trees (ET), K-Nearest Neighbor (KNN), and Adaptive Boosting (AdaBoost) can accurately predict the survival rate of the tested samples, which is 75,4\%, 74,7\%, 71,5\%, 75,5\%, 70,3\%, and 78 percent.
\end{abstract}


\section{Introduction}
For more than four decades, the collection, intersection and analysis of massive data have been fundamental issues in the health sciences, requiring the use of artificial intelligence concepts, techniques, and tools to process, optimize and improve care by helping health professionals improve their efficiency, productivity, and consistency in the quality of care provided to patients \cite{yu2018artificial, harnoune2021bert}. Although it is mostly used in fields such as endocrinology-nutrition and hepato-gastroenterology, it now touches all medical specialties, including epidemiology, which is defined as the study of the relationship between diseases and the factors that may cause or influence their frequency, spatial distribution and evolution. 

Many epidemiologists are looking to incorporate artificial intelligence (AI) tools and techniques into their research, as it can have many impacts in analytical research. AI can be defined as the ability of a machine to learn and recognize patterns and relationships from a sufficient number of representative examples and to use this information effectively to make decisions on unseen data \cite{thiebaut2018artificial}. For this reason, the use of AI methods has become essential in the field of epidemiology. It can find correlations between certain behaviors or socio-demographic characteristics and the presence of diseases, such as cancer, it can also predict the prevalence of infectious diseases and the survival time of a patient by considering many interdependent factors. And as in other areas of medicine, AI plays a role in detecting and mapping diseases \cite{Mikram2021, abdoul2021hybrid}, especially those closely related to environment and behavior, such as cancer.

In this paper, we concentrate on breast cancer, Many of us know someone who is struggling with breast cancer, or at least hear about the challenges faced by patients battling this cancer. Breast cancer (BC) is a cancerous tumor that originates in the breast tissue. It can spread straight to surrounding areas or to distant parts of the body. The cancer occurs almost exclusively in females, but males can also develop this type of cancer \cite{bib2}. The malignant tumor is a group of cancerous cells that can spread to nearby tissues and destroy them. It can also extend to other parts of the body. It is the most common cancer diagnosed and one of the main causes of death in the female population. More than 2.26 million new breast cancer cases have been reported in women \cite{bib3}. In 2020, 2.3 million women in the world were diagnosed with breast cancer and 685,000 died from it. By the end of 2020, 7.8 million women alive had been diagnosed with breast cancer in the preceding five years, which makes it the most common cancer in the world \cite{bib4}. Based on the most recent GLOBOCAN [World Health Organization 2020] statistics, it is the number one commonly diagnosed cancer and the number five reason for cancer deaths in the world, accounting for 6.84\% of all mortalities \cite{bib5}.

Many analytical investigations have predicted breast cancer survival using machine learning techniques in order to reduce the negative effects of BC on human health. In order to forecast the survival of patients with a specific disease over time, a variety of methodologies are used to data that are recorded in health datasets. This process is known as survival analytics. These techniques include, for instance, machine learning (ML) models \cite{bib6}. A branch of artificial intelligence known as "machine learning" employs methods that let computers draw on past performance to get better at what they do. Malignant and benign cancers may be distinguished using machine learning algorithms, and breast cancer survival can also be predicted \cite{bib7}. In general, these methods enable the creation of adaptive systems from different data sets, the discovery of latent links between data components, and the prediction of events\cite{bib8}.

The topic of this article is clinical survival prediction. We used METABRIC (Molecular Taxonomy of Breast Cancer International Consortium), which analyzes the patterns of molecules within tumors of close to two thousand women, for whom information about tumor characteristics had been meticulously recorded, to test the relationship between clinical features and outcomes. An exploratory data analysis must initially be performed on the METABRIC dataset. Risk factors are found using various techniques in the second stage. The precision, false positive rate, true positive rate, accuracy, recall, and confusion matrix are the metrics used to compare the performance of each classification method in the third stage.

The rest of the text is structured as follows: The contributions of artificial intelligence to epidemiology are discussed in Section 2, along with some examples of how it has been used in this discipline. The many approaches employed by various authors to forecast breast cancer survival are covered in Section 3, with particular emphasis on the datasets, factors included in the research under review, and machine learning algorithms. More information about the procedures employed in our investigation is provided in Section 4, and the findings are presented in Section 5.

\section{Artificial Intelligence in Epidemiology}\label{ep}
The study of diseases' relationships to potential causes or factors that may affect their incidence, spatial distribution, and evolutionary trends is known as epidemiology. It has a crucial role in public health strategies, particularly in recognizing or avoiding the emergence of new diseases or the recurrence of old ones. Only recently, in 1854, was the topic of epidemiology acknowledged as a legitimate academic discipline. Yet it is a cornerstone of both general care and preventive medicine in particular. Probability theories and statistical techniques have traditionally been used in epidemiology. In this way, it is one of the fields that has long utilized what is now known as big data. Artificial intelligence can be used in a variety of ways in this area of research to define the environmental causes of diseases through analytical studies. For instance, it can be used to determine links between particular habits or sociodemographic details and the emergence of ailments like diabetes or cancer. \cite{bib1}. 

Cancer is just one application of artificial intelligence in epidemiology. A similar connection between the environment and obesity rates was discovered in 2018 by a team of Americans from the University of Washington lead by Adyasha Maharana. \cite{bib9}. They conducted their study in two steps: first, they retrieved built-environment features from satellite data using a Convolutional Neural Network (CNN), and then they extracted and analyzed Point of Interest (POI) Data. They next developed a compact model utilizing net elasticity regression to evaluate the relationship between the built environment and the prevalence of obesity. As a result, various diseases that are closely tied to the environment can be found using this technique. The term "infectious disorders" immediately brings to mind COVID-19 (Coronavirus Disease 2019). The same method can be used by Deep Learning to comprehend the pandemic's evolution rates, forecast them, and modify health precautions. A team from the University of Tokyo predicted the spatiotemporal distribution of dengue in Taiwan in 2019 using data on sea temperature and rainfall \cite{bib10}. We can battle infectious diseases like the Zika virus, dengue, or chikungunya based on the hypothesis that artificial intelligence can be used to model the distribution of mosquito nests\cite{bib11}.

\section{Related Works}\label{rw}

In order to identify the patterns associated with breast cancer patient survival, a number of machine learning techniques have already been employed to predict breast cancer survival, including logistic regression, KNNs, Bayesian networks, support vector machines, and decision trees. Many prognostic factors, including age, race, marital status, primary site, laterality, behavioral code, histology, tumor size, lymph node, extension, surgery, radiation, and TNM stage, were chosen in each of these studies. However, the handling of unknown or missing values is another issue that has been handled differently by various studies.

By using machine learning and deep learning approaches, Kalafi et al.\cite{bib7} show a modest improvement in the accuracy of breast cancer survivability prediction. Their research finds the most crucial survival indicators (tumor size, age, total number of axillary lymph nodes removed, stage, and number of positive nodes). Then, they assessed the sensitivity, specificity, accuracy, F1 score, negative predictive value, false positive rate, false discovery rate, false negative rate, and Matthew correlation coefficient of the prediction models of Random Forest, Decision Trees, Support Vector Machine, and Multilayer Perceptron.

For the purpose of predicting survival rates for various subtypes of breast cancer, Montazeri et al.\cite{bib12} describe a rule-based classification strategy that makes use of machine learning techniques. They used a dataset with eight attributes that contained the records of 900 patients, 876 (97.3`\%) of whom were female, and 24 (2.7\%) of whom were male. They employed Naive Bayes, AdaBoost, Tree Random Forest, 1-Nearest Neighbor, Support Vector Machine, RBF Network, and Multilayer Perceptron in their research (MLP). Precision, sensitivity, accuracy, specificity, and AUC were used to gauge how well each ML approach performed ( Area Under the ROC Curve).

Machine learning techniques were used by Ganggayah et al.\cite{bib13} to examine breast cancer survival predictive variables. The random forest algorithm's model evaluation accuracy was a little bit better than that of the other algorithms. However, the level of accuracy demonstrated by each method appeared to be manageable. Their study found six factors to be particularly important: the cancer stage, the size of the tumor, the total number of axillary lymph nodes excised, the  presence of positive lymph nodes, the technique of diagnosis, and the primary therapy type. The process of selecting variables in healthcare research, particularly when using ML approaches, may produce diverse findings based on the dataset, geography, and patient lifestyle.This study identifies the model's performance as well as the crucial factors influencing breast cancer patients' survival that may be applied in clinical practice. This study demonstrates how visualization of outcomes can be utilized to create predictive survival applications by building decision trees and survival graphs to enable validation of the major variables determining breast cancer survival.

\section{Methods}\label{sec4}
On the METABRIC (Molecular Taxonomy of Breast Cancer International Consortium) dataset, the models were developed and evaluated. This dataset for 1904 breast cancer patients comprises 175 gene mutations, 31 clinical characteristics, and the mRNA z-scores for 331 genes. In the dataset, we solely kept the clinical variables. In order to determine whether the cancer patients would survive, we conducted analysis and created a model. There are a number of algorithms that can be applied to this binary classification problem. We evaluated these algorithms with the default settings to obtain a general feel of how well they performed before doing 10-fold cross-validation for each test because we are unsure yet whether one will perform better. \cite{bib14}.

\subsection{K-fold cross validation strategy}\label{subsec4}
The 10-fold cross-validation approach, which separates the data into k subgroups, was used in this paper. Each of these subgroups is used for validation and k-1 is used for training in every k iterations. All data will be used exactly k times for training and once for testing as this technique is performed k times. The last estimated value is the average of the k-fold validation results. \cite{bib15}. 

\subsection{Logistic Regression Classification}\label{subsec5}

One technique that could be utilized to address binary classification issues is logistic regression (LR). This approach is based on the sigmoid function, an S-shaped curve that can be used to give a real number a value between 0 and 1, but never precisely within these bounds. When it is desired to be able to forecast whether cancer patients will survive based on the values of a number of predictor factors, LR is utilized. Despite being comparable to a linear regression model, it is suitable for models where the binary variable of interest\cite{bib16} \cite{bib17}. 

The LR model for F independent variables can be expressed as follows :
\[
    F(X = 1)  =  \frac{1}{1+e^{\beta_0 + \beta_1y_1 +  \beta_2y_2+...+ \beta_py_p}}
    \\
\]
with F(X = 1) being the chance of survival of the cancer patient, and the regression coefficients are $\beta_0,\beta_1,...,\beta_p$. 

There exists a hidden linear model in the LR model. The natural logarithm of the ratio of F(X = 1) = (1- F(X = 1) leads to a linear model in $y_i$ :

$p(y) = \ln\frac{F( X = 1 )}{1 - F( X = 1)}
    = \beta_0 + \beta_1y_1 +  \beta_2y_2+...+ \beta_py_p$

Many of the necessary characteristics of a linear regression model are present in the p(y). The independent variables may include mixtures of categorical and continuous variables \cite{bib18}\cite{bib19}.

\subsection{Support Vector Machines (SVM) Classification}\label{subsec6}

In recent years, support vector machines (SVMs) have advanced quickly. The SVM learning problem is the term used to describe an uncertain and nonlinear dependence between a high-dimensional input vector x and a scalar output y. It is significant to note that no knowledge of the joint probability functions is accessible, necessitating the use of free distribution learning. The only information that is provided is a set of training data T = ($x i$,$y i$) $\in$ X Y,i=1,l, where l denotes the number of training data pairs and is thus equal to the size of the training data set D. Additionally, $y i$ is marked as $d i$, where d denotes a desired (target) value. Consequently, SVMs are a component of supervised learning approaches.\cite{bib20}\cite{bib21}.

\subsection{Decision Tree Classification}\label{subsec11}

A decision tree (DT) is a form of tree structure resembling a diagram in which each leaf node denotes a result, each branch denotes a decision rule, and each internode denotes a feature. A DT is used to examine a piece of data in order to create a set of guidelines or inquiries that are used to forecast a class. By learning basic decision rules from the features in the data, a DT aims to create a model that can predict the value of a target variable. In this way, a DT selects the best attribute to divide the records, turns that attribute into a decision node, divides the data set into smaller subsets, and repeats this process iteratively until it starts to construct a tree.\cite{bib23}.
 
\subsection{Random Forest Classification}\label{subsec12}

The supervised learning method known as random forests (RF) enables us to create multiple predictors before combining their various predictions to produce categorization predictive models. A group of decision trees that were typically trained using the bagging method make up these forests. The fundamental goal of bagging is to build an average of numerous noisy but roughly unbiased models in order to reduce variation\cite{bib25}. The following algorithm is used to construct each tree :

\begin{itemize}
\item Consider X as the number of testing cases, Y as the number of variables in the classifier.
\item Consider z as being the number of input variables to be applied to identify the decision in a particular node, z $\prec$ Y.
\item Pick a training set for the tree and use the rest of the test cases to approximate the error.
\item At each node of the tree, pick randomly z variables on which the decision should be based. Calculate the best training set score from the z variables.
\end{itemize}

A new case gets lowered in the prediction tree. The terminating node is then shown on the label. The process is really repeated for each tree in the forest, and the prediction is given for the tag with the greatest number of occurrences. We state that there are 100 trees in the forest as a whole \cite{bib25} \cite{bib24}.

\subsection{Extremely Randomized Trees Classification}\label{subsec13}

A machine learning approach called Extremely Randomized Trees or Extra Trees combines the predictions of various decision trees. It is comparable to the common RF algorithm. Despite using a simpler approach to create the decision trees that make up the ensemble, it frequently achieves performance that is comparable to or better than the RF algorithm\cite{bib26}.

\subsection{K-Nearest Neighbor Classification}\label{subsec14}
K-Nearest Neighbors (KNN) is a technique for keeping track of all examples that are accessible and evaluating new cases based on a similarity metric \cite{bib27}. This approach is non-parametric because it doesn't assume anything about the distribution of the data at its core, and it's also less labor-intensive because it doesn't call for building a model from training data. The test stage makes use of all the training data. As a result, testing is more expensive and takes longer than training. In this strategy, the number of neighbors k is typically uneven if the total number of classes is 2. Distances like the Euclidean, Hamming, Manhattan, and Minkowski must be calculated to identify the closest spots \cite{bib28}.

\subsection{Adaptive Boosting Classification}\label{subsec15}

One of the boosting ensemble classifiers that Yoav Freund and Robert Schapire first presented in 1996 is adaptive boosting, also known as Ada-Boost. In order To strong and accurate classifier, it combines several weak classifiers \cite{bib29}. It operates in aby following steps:

Adaboost first selects a subset of training at random, then iteratively trains the model by selecting the training set based on the recent training's accurate predictions. Next, it assigns the highest weight to the misclassified observations so that they will have the highest probability of being classified in the upcoming iteration. It also assigns the highest weight to the trained classifier in each iteration based on the classifier's precision. The classifier with the highest accuracy will be given more weight. Finally, this is iterated until all training data matches accurately throughout or until the maximum number of specified estimators are reached, whichever comes first\cite{bib30} The process is depicted in Fig. 1.

\begin{figure}
\centering
\includegraphics[width=.8\linewidth]{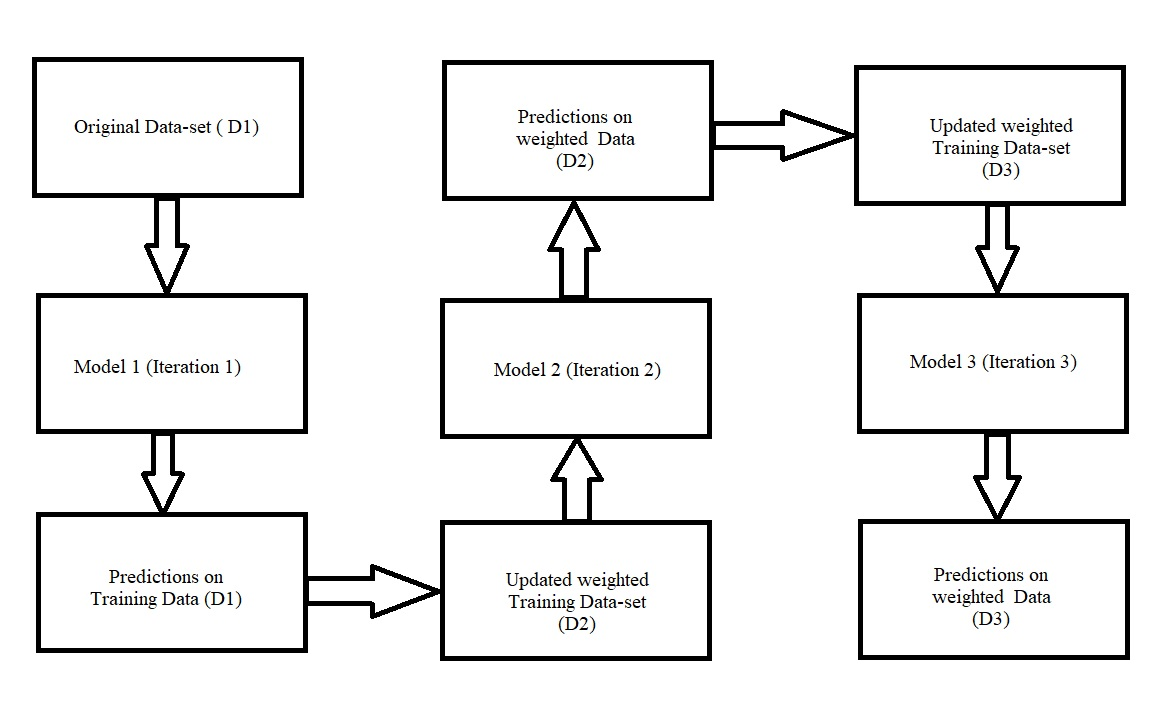}
\caption{The Ada Boost algorithm's steps}
\label{fig1:computerNo}
\end{figure}
\
\section{Results }\label{sec}

\subsection{Data Description}\label{data}

A Canadian-British collaboration produced the METABRIC (Molecular Taxonomy of Breast Cancer International Consortium) dataset, which includes chosen sequencing information from 1,980 primary breast cancer samples. The cBioPortal was used to upload the clinical and genetic data. Professor Sam Aparicio of the British Columbia Cancer Centre and Professor Carlos Caldas of the Cambridge Research Institute in Canada put together the dataset, which was then published in Nature Communications (Pereira et al., 2016). 1904 patients made up the data set for this investigation, of which 57.9\% were still alive and 42.1\% had passed away\cite{bib31}.

\subsection{Performance of the proposed methods}\label{res}
Many variables are involved in some predictive modeling issues, which can slow down model development and learning and demand a lot of system memory. Additionally, some models' performance may worsen when input variables that are unrelated to the target variable are used. Because of this, many data analysts employ techniques that enable them to find the most crucial features. In our analysis, we employed SelectKbest, which calculates the relative relevance of each variable based on the highest correlation between the variables and survival time. According to Fig. 2, which displays the importance scores for each variable included in this analysis, TNM, age at diagnosis, cohort, and tumor stage were, in that order, the variables most important for predicting breast cancer survival.
The recommended ML models are applied to forecast the prognosis of breast cancer patients. Seven distinct models : LR, SVM, DT, RF, T, KNN, and Adaboost, were used in this study. Precision, false positive rate, true positive rate, recall, accuracy, AUC, and confusion matrix were just a few of the metrics used to gauge their effectiveness. When applying 10-fold ccross-validationon on each model, the results provided in Table 1 and the box-plots in Fig. 3 demonstrate the superior performance of Adaboost. By contrasting the rates of true positives and the rates of false positives for various threshold levels, the ROC curve (Receiver Operating Characteristic) shown in Fig. 4 enables one to assess how well the algorithms we used in our study performed. It demonstrates that the AdaBoost algorithm is the most predictable algorithm.

The prediction level and significance test results for the survival prediction in LR, SVM, DT, RF, ET, KNN, and Adaboost are shown in Figure 5. The confusion matrix reveals that the accuracy of LR, SVM, DT, RF, ET, KNN, and Adaboost is 75.4\%, 74.1\%, 71.1\%, 75.1\%, 70.1\%, 73.1\%, and 78.1\%, respectively. Adaboost had the highest accuracy for survival prediction in breast cancer using the METABRIC dataset, and Extra trees had the lowest prediction accuracy of all classifiers. Adaboost achieved the best performance, dominating all evaluation metrics as shown in Table 1, Fig. 3, and Fig. 4.

\begin{figure}
\centering
\includegraphics[width=0.8\linewidth]{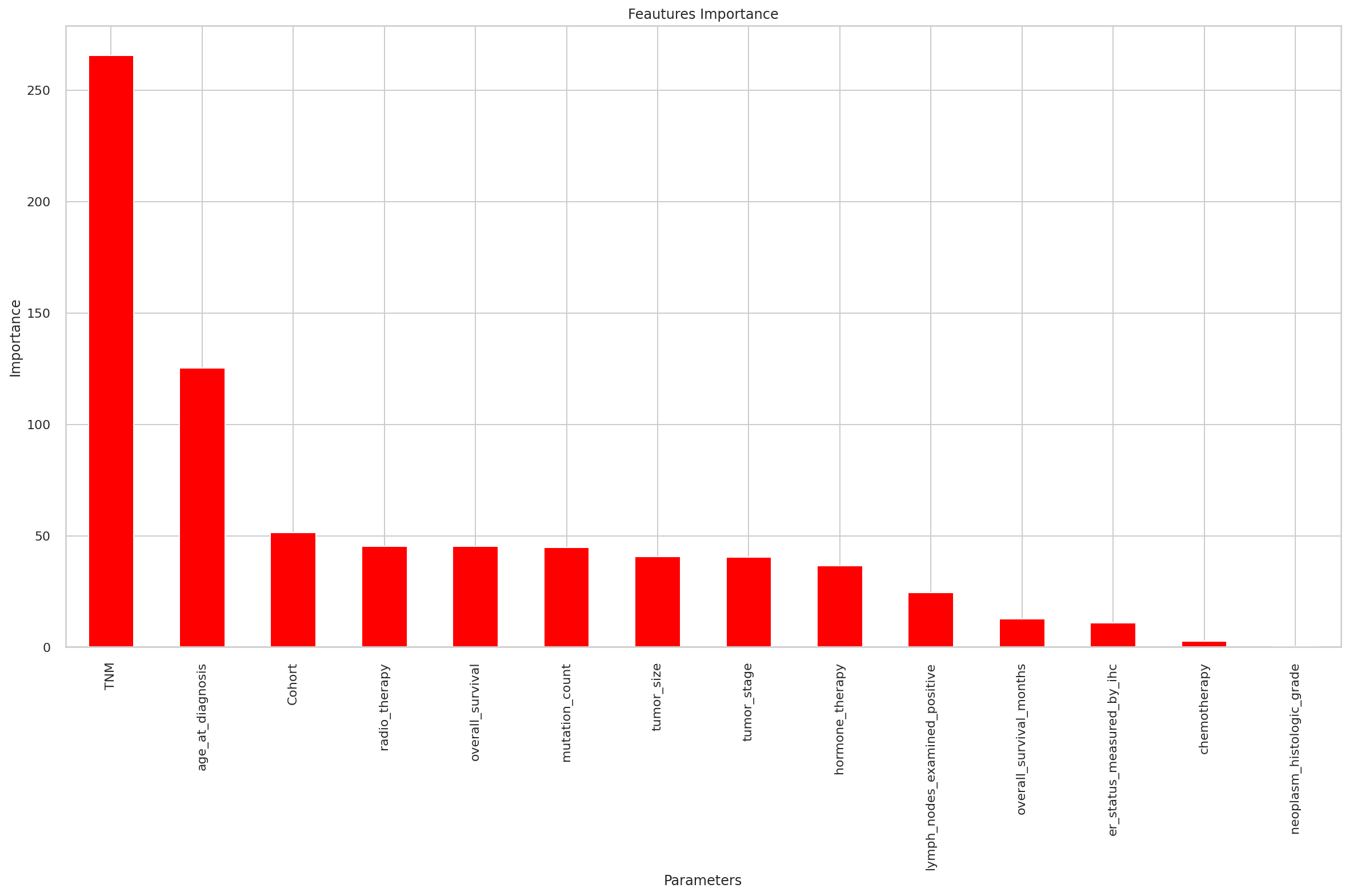}
\caption{The significance of many factors in determining breast cancer survival}
\label{fig:computerNo1}
\end{figure}
\
\begin{table}[]
\centering
\resizebox{\columnwidth}{!}{%
\begin{tabular}{|c|l|l|l|l|l|}
\hline
MLA used            & Train Accuracy & Test Accuracy & Precision & Recall    & AUC       \\ \hline
Logistic Regression & 0.7754         & 0.7731        & 0.743169  & 0.727273  & 0.767718  \\ \hline
AdaBoost            & 0.8244         & 0.7870        & 0.771429  & 0.721925 & 0.779330  \\ \hline
Extra Trees         & 1.0000         & 0.7153        & 0.716216  & 0.566845  & 0.697708  \\ \hline
Random Forest       & 1.0000         & 0.7778        & 0.775758  & 0.684492  & 0.766736  \\ \hline
SVM                 & 0.7127         & 0.7361        & 0.721212  & 0.636634  & 0.724304  \\ \hline
Decision Tree       & 1.0000         & 0.6875        & 0.644444  & 0.620321  & 0.6724304 \\ \hline
K-Neighbors         & 0.8107         & 0.6875        & 0.639785  & 0.636364  & 0.681447  \\ \hline
\end{tabular}%
}
\caption{A comparison of all machine learning algorithms}
\end{table}

\begin{figure}
\centering
\includegraphics[width=.8\linewidth]{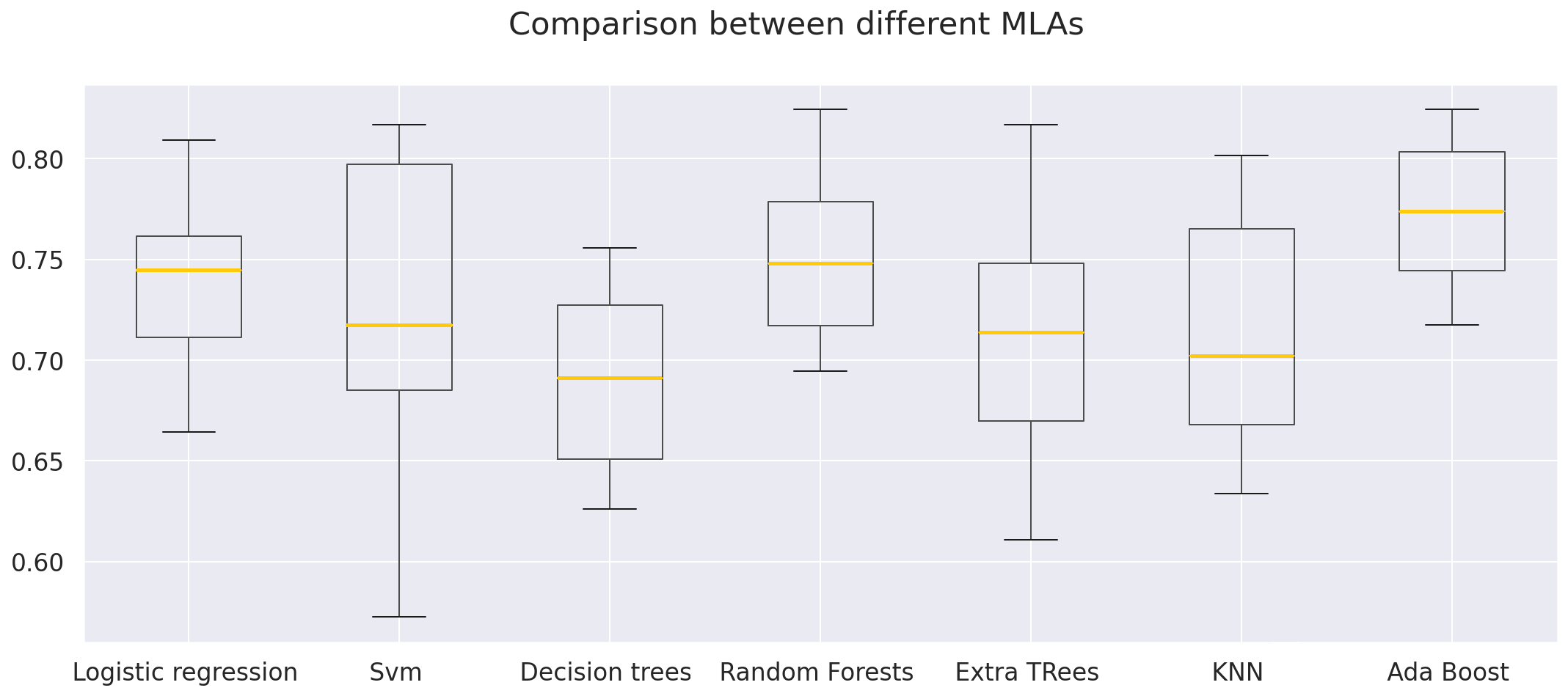}
\caption{Performance of LR, SVM, DT, RF, ET, KNN, and Adaboost based on accuracy with 10-fold cross validation}
\label{fig:computerNo2}
\end{figure}
\

\begin{figure}
\centering
\includegraphics[width=.8\linewidth]{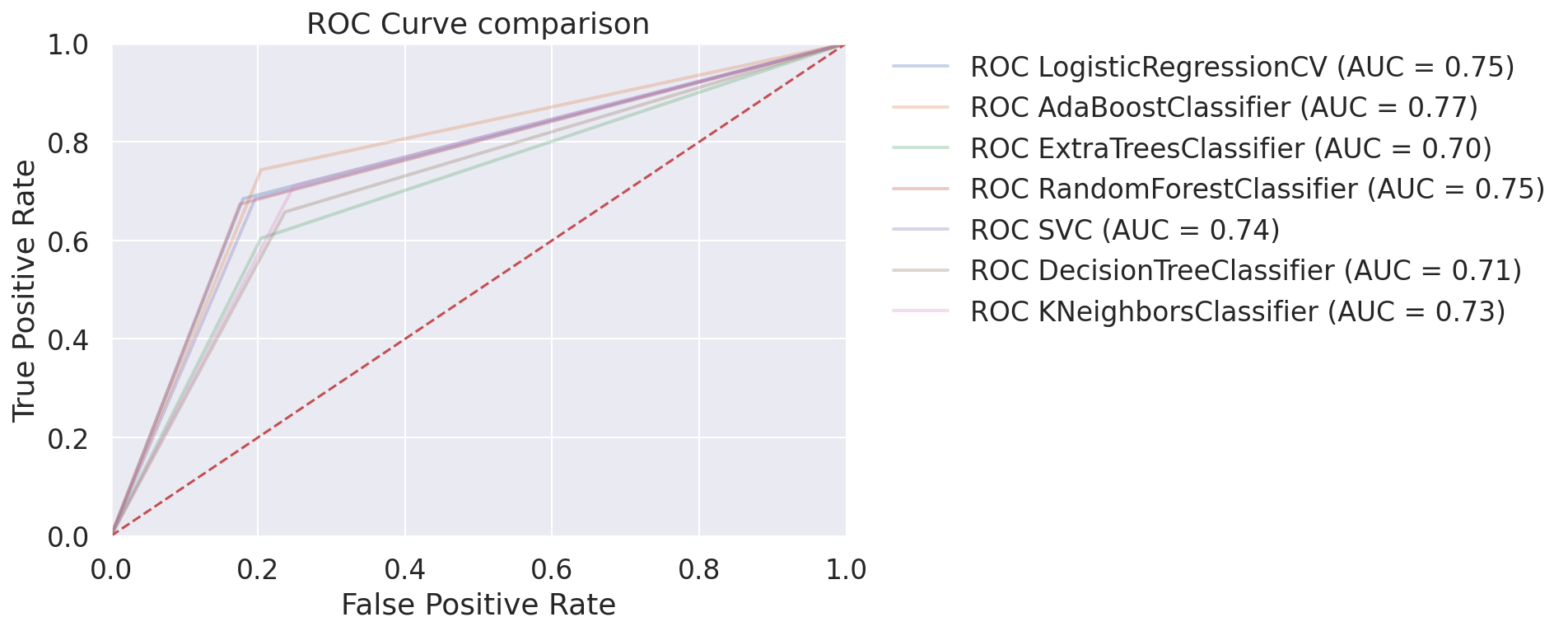}
\caption{Roc Curve Comparaison}
\label{fig:computerNo3}
\end{figure}
\
\begin{figure}
\centering
\includegraphics[width=.8\linewidth]{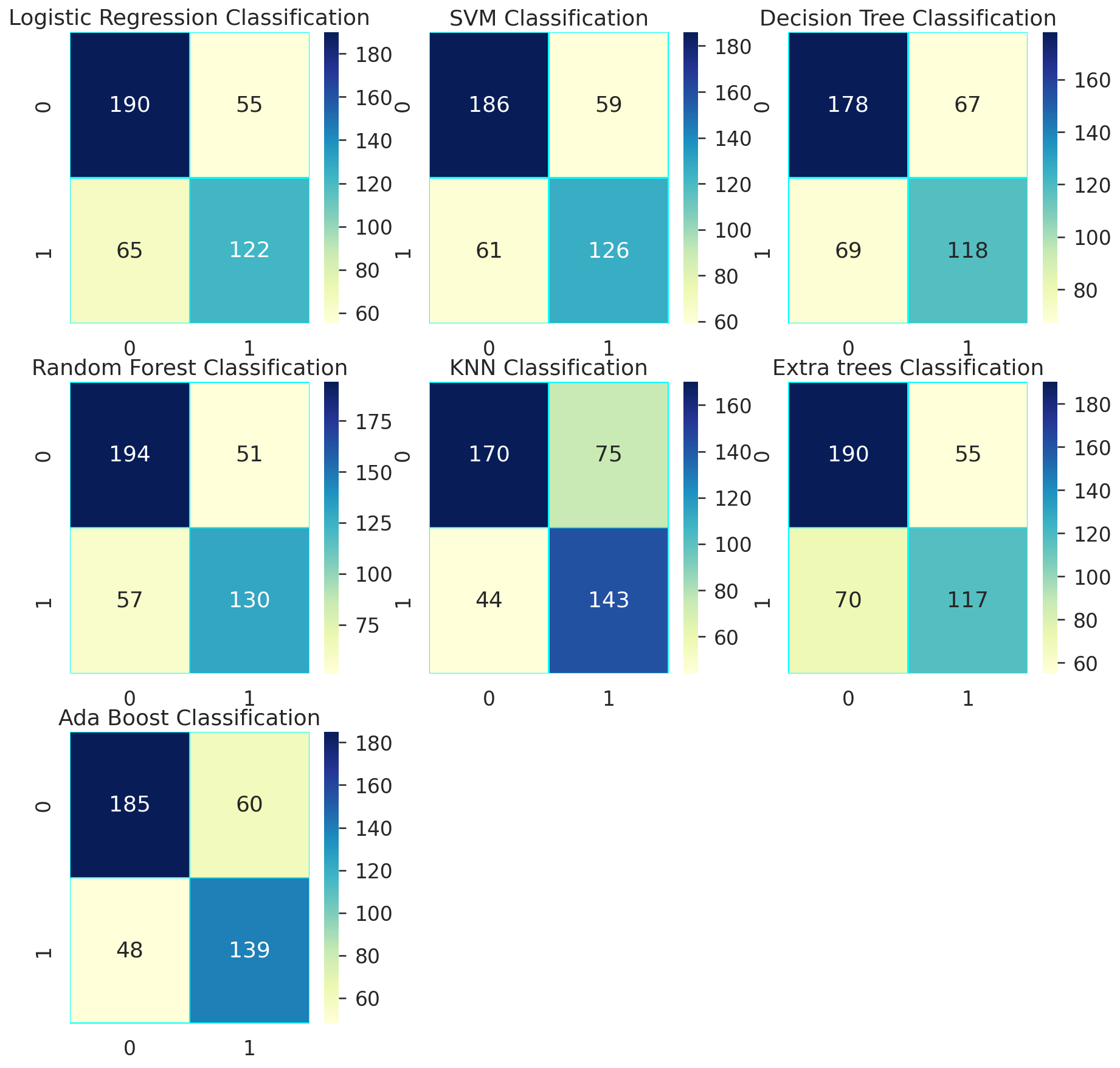}
\caption{Confusion matrix of LR, SVM, DT, RF, ET, KNN and Adaboost results}
\label{fig:computerNo}
\end{figure}
\

\section{Conclusion And Recommendation}\label{sec6}
The proper diagnosis of breast cancer is crucial to save the lives of numerous people. Researchers are currently very interested in using machine learning classification algorithms to predict whether a cancer patient will survive, regardless of the usage of current diagnostic tools. With regard to machine learning classification approaches, such as LR, SVM, DT, RF, ET, KNN, and Ada Boost techniques, this study was conducted to assess their performance.These methods work well as survival prediction tools. With an accuracy of 78\%, we discovered that the Ada Boost algorithm produced the most accurate results.

By gathering more data or creating new data that are statistically "near" to the existing data, we can enhance the performance of the models employed to better support the study's findings.

\bibliographystyle{unsrt}  

\bibliography{references}

\end{document}